\documentclass[cameraready]{Interspeech}

\title{Speaker-Specific and Language-Dependent Temporal Organization in Bilingual Political Speech}

\author[
    affiliation={1,2},
    orcid=0000-0002-0821-9125,
    correspondingauthor
]{Nina}{Hosseini-Kivanani}

\author[
    affiliation={3},
    orcid=0000-0001-9715-4505,
]{Nafiseh}{Taghva}

\author[
    affiliation={1},
    orcid=0000-0001-9216-3082
]{Peter}{Gilles}

\author[
    affiliation={4},
    orcid=0000-0002-8623-1680
]{Oliver}{Niebuhr}

\address{$^1$ Faculty of Humanities, Education and Social Sciences, University of Luxembourg \\
    $^2$Radio Télévision Luxembourg (RTL), Luxembourg\\
    $^3$ Department of Foreign Languages and Linguistics,
    Shiraz University, Iran  \\
    $^4$ Centre for Industrial Electronics, University of Southern Denmark, Denmark 
}

\email{nina.hosseinikivanani@ext.uni.lu,taghvanafiseh@gmail.com,peter.gilles@uni.lu,olni@sdu.dk
}


\usepackage{comment}
\usepackage[utf8]{inputenc}
\usepackage[table]{xcolor} 
\usepackage{booktabs}

\usepackage{graphicx}      
\usepackage{array}         
\usepackage{amssymb} 
\usepackage{textgreek} 
\usepackage{newunicodechar}
\usepackage{stackengine} 
\newunicodechar{≥}{\geq}
\newunicodechar{−}{-}
\newcolumntype{C}{>{\centering\arraybackslash}m{0.095\textwidth}}
\usepackage{tabularx}

\definecolor{luxblue}{RGB}{0,114,178}   
\definecolor{frorange}{RGB}{230,159,0}  

\newcommand{\hlLux}[1]{\cellcolor{luxblue!12}{#1}}
\newcommand{\hlFr}[1]{\cellcolor{frorange!12}{#1}}

\begin{document}

\maketitle

\begin{abstract}

Speech rhythm helps structure persuasive speech, but most empirical work examines monolingual English. This study asks how politicians organize timing when speaking Luxembourgish and French. We analyze 400 sentences from ten politicians, annotated for segments and pauses. We compute rhythm metrics, including means, variability, and pairwise variability indices for consonants and vowels. We quantify speaker and language contributions and test within-speaker language effects with paired t-tests. Results show that consonant-based metrics retain speaker-specific signatures, whereas vowel-based metrics are largely driven by language choice. French tokens display longer and more variable vowels and vocalic intervals, while consonant timing differences are smaller. No robust language by gender interactions emerge. These findings show that language choice systematically reorganizes rhythmic timing in bilingual public speech. 

\end{abstract}

\section{Introduction}

Timing and rhythm are central to how speakers structure messages and manage listener attention. Research on persuasive and charismatic public speaking shows that successful speakers adjust not only pitch and loudness but also pauses, segment timing, and phrase-level tempo contours in systematic ways~\cite{niebuhr2024rhythm,bosker2020contribution,mixdorff2018model,niebuhr2018shapes}. Speech rhythm research has moved from categorical stress-timed versus syllable-timed typologies~\cite{abercrombie2019elements,pike1945intonation} toward continuous, multidimensional acoustic approaches~\cite{grabe2026durational,ramus1999correlates}. Duration-based metrics such as the proportion of vocalic intervals
(\%V), consonantal variability ($\Delta C$), and raw and normalized
pairwise variability indices (rPVI, nPVI) are standard tools for
quantifying rhythmic patterns across languages and speakers
~\cite{arvaniti2012usefulness,dellwo2010influences,white2007calibrating}.
Although they capture only selected aspects of durational organization rather than rhythm as a perceptual construct~\cite{white2020speech}, these metrics are useful for comparing speaking styles, language varieties, and communicative functions including emphasis, fluency, and affective stance~\cite{keller2007speech}.

Most existing work combines monolingual speakers with controlled tasks. Niebuhr and Taghva~\cite{niebuhr2024rhythm}, for example, analyzed English investor pitches in neutral and charismatic styles and showed that charismatic performances display greater durational variability in consonantal and voiced intervals, with increased raw PVI, while overall speaking rate remains constant. Listener charisma ratings were most strongly linked to consonantal and voiced interval variability. These findings indicate that rhythm metrics capture meaningful within-language style shifts. At the same time, recent corpus work suggests that some metrics reflect stable speaker signatures while others capture language-inherent timing. Taghva et al.~\cite{taghva2023corpus} showed that \%V and syllable rate effectively index between-speaker differences in Kalhori Kurdish across read and spontaneous speech, highlighting the need to separate language-specific from speaker-specific contributions in bilingual settings.

Cross-linguistic influence on bilingual rhythm is widely documented. Taghva and Chaudhuri~\cite{taghva2024language} found that Bengali advanced learners of English retain a faster tempo and higher \%V in their L1, while their L2 shows greater consonantal variability ($\Delta C$, nPVI C), with sonority patterns remaining comparable across languages. Similar results for Indian English~\cite{fuchs2016speech} and Galician learners~\cite{rodriguez2023l2} show that L1 rhythmic properties permeate L2 production, especially at lower proficiency levels. These studies provide methodological evidence for separating speaker-, style-, and language-related timing effects, rather than direct typological parallels.
This aligns with work on a “charisma gender gap”, where women often receive lower charisma ratings despite similar melodic patterns and may require stronger or more consistent acoustic cues, for instance through filled pauses and prosodic emphasis~\cite{novak2017gender,voss2024beautiful}.

\begin{table*}[hpt!]
  \centering
  \scriptsize
  \setlength{\tabcolsep}{0pt}        
  \renewcommand{\arraystretch}{0.9}  

  \caption{Selected Luxembourgish politicians and their current roles.}
  \label{tab:lux_politicians}

  \begin{tabular}{@{}*{10}{C}@{}}
    \includegraphics[width=\linewidth,height=1.5cm]{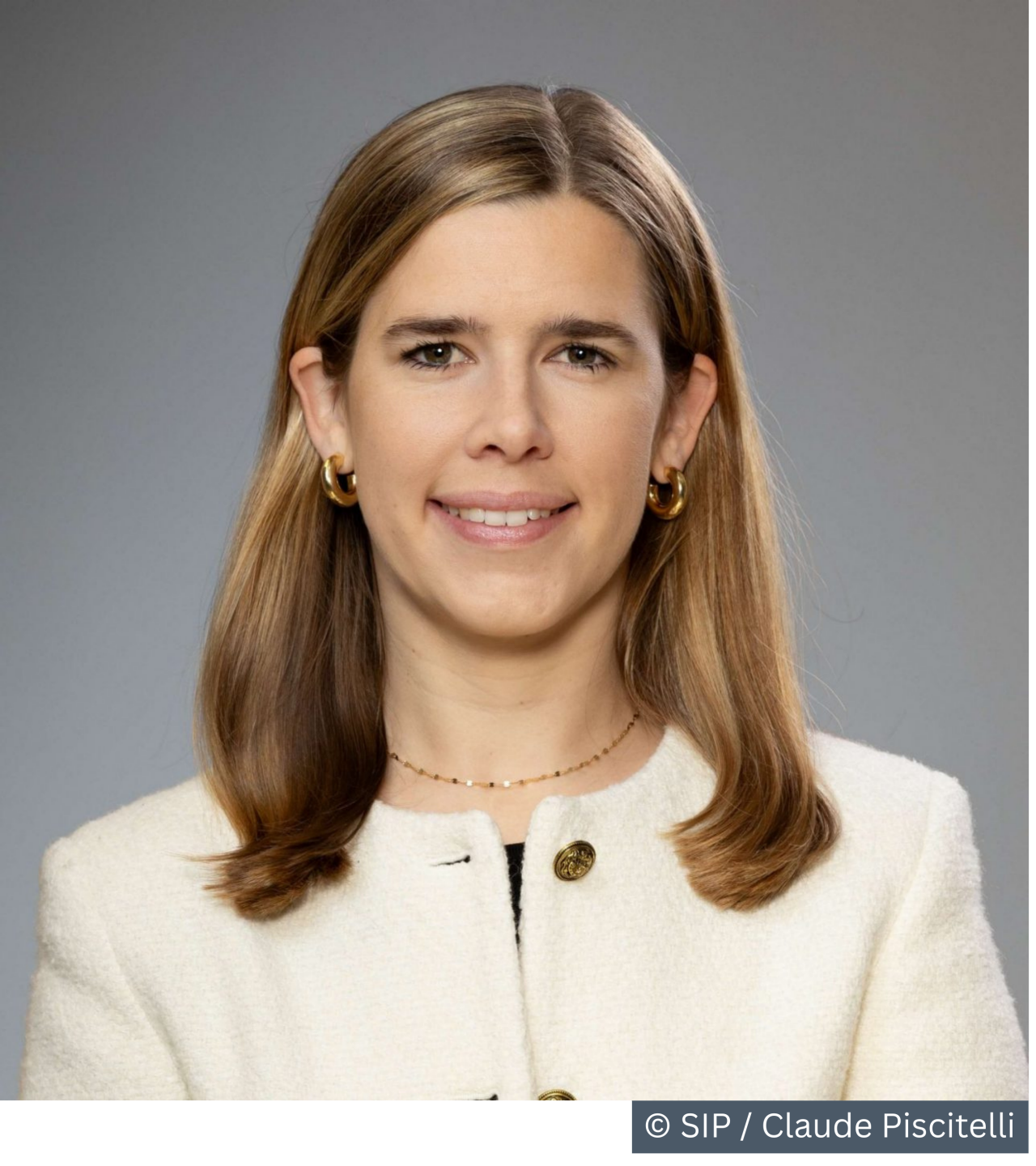} &
    \includegraphics[width=\linewidth,height=1.5cm]{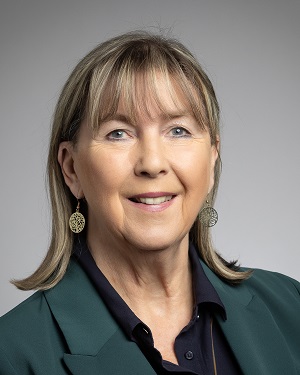} &
    \includegraphics[width=\linewidth,height=1.5cm]{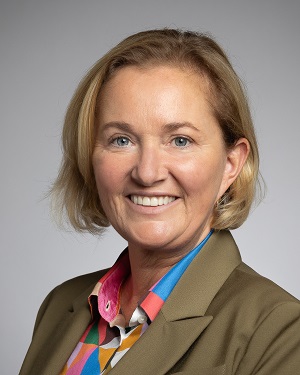} &
    \includegraphics[width=\linewidth,height=1.5cm]{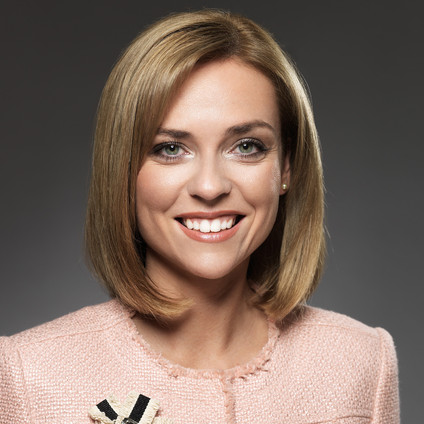} &
    \includegraphics[width=\linewidth,height=1.5cm]{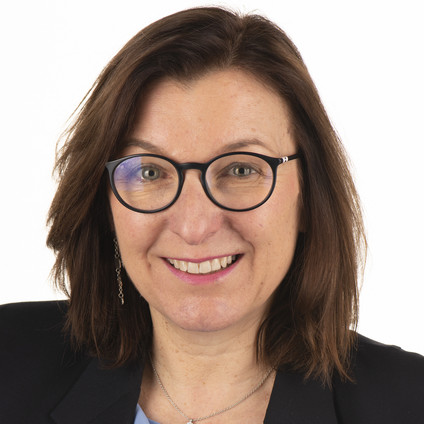} &
    \includegraphics[width=\linewidth,height=1.5cm]{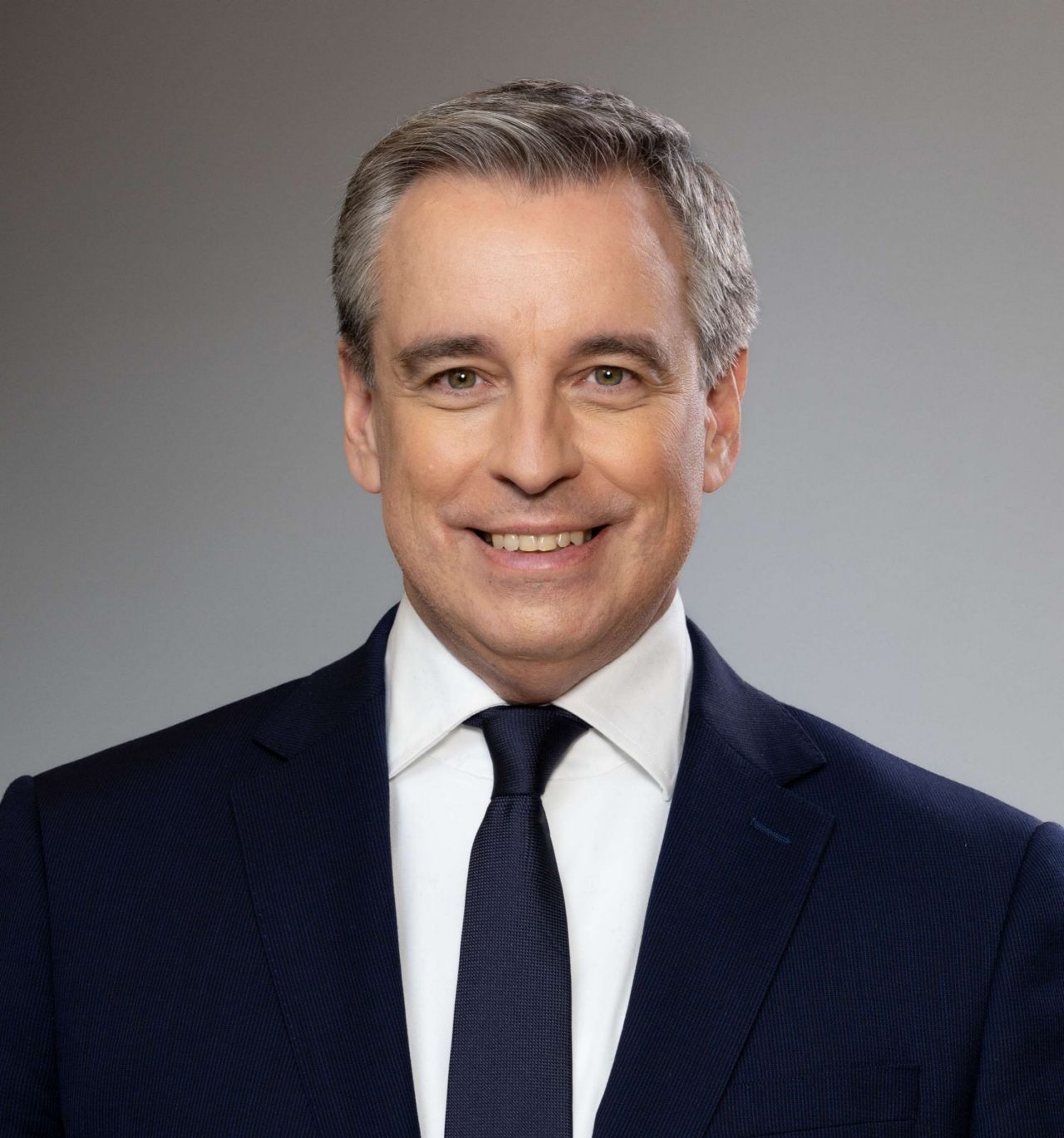} &
    \includegraphics[width=\linewidth,height=1.5cm]{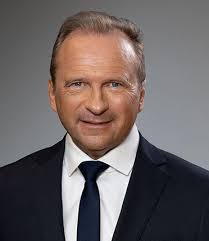} &
    \includegraphics[width=\linewidth,height=1.5cm]{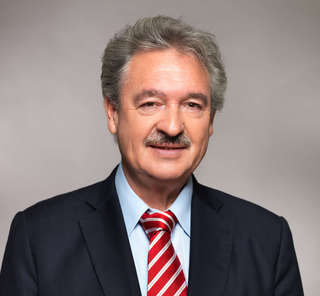} &
    \includegraphics[width=\linewidth,height=1.5cm]{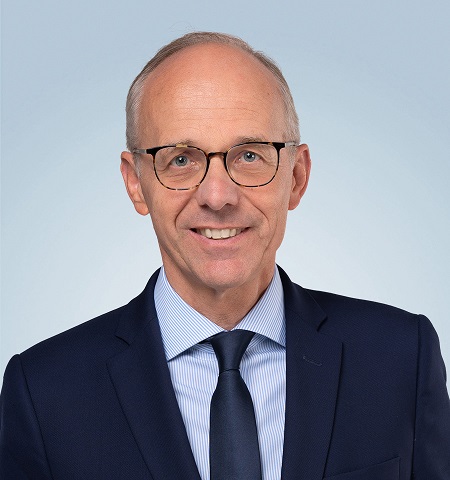} &
    \includegraphics[width=\linewidth,height=1.5cm]{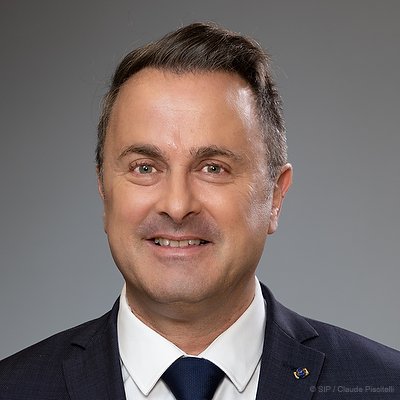} \\[-2pt]
    {\tiny\shortstack{Elisabeth\\Margue, 35\\Justice minister;\\Del. PM (media,\\connectivity, parliament)}} &
    {\tiny\shortstack{Lydie\\Polfer, 73\\Mayor of\\Luxembourg City}} &
    {\tiny\shortstack{Paulette\\Lenert, 57\\MP (LSAP); former\\Dep. PM, health}} &
    {\tiny\shortstack{Taina\\Bofferding, 43\\MP (LSAP);\\Socialist group\\president}} &
    {\tiny\shortstack{Tilly\\Metz, 58\\MEP (Greens/EFA,\\déi Gréng)}} &
    {\tiny\shortstack{Claude\\Meisch, 54\\Education, children\\and youth; housing\\and planning}} &
    {\tiny\shortstack{Gilles\\Roth, 58\\Minister of\\Finance}} &
    {\tiny\shortstack{Jean\\Asselborn, 76\\Former foreign and\\European affairs\\minister}} &
    {\tiny\shortstack{Luc\\Frieden, 62\\Prime Minister;\\CSV president}} &
    {\tiny\shortstack{Xavier\\Bettel, 52\\Dep. PM; foreign\\affairs and trade;\\development coop.}} \\
  \end{tabular}
\end{table*}

How speakers organize rhythm across languages in public speech, where language choice carries social meaning, remains understudied. Luxembourg offers a clear case: Luxembourgish indexes everyday interaction and national identity, whereas French dominates administration, law, and other prestige domains~\cite{horner2016language,gilles2013luxembourgish}. Politicians and public figures use both languages in the same media environment and often address similar audiences in formal public settings. Political speech is a useful test case because bilingual speakers must satisfy language-specific timing constraints while maintaining audience-oriented authority, clarity, and presence~\cite{niebuhr2018shapes,rosenberg2009charisma}. Yet no study has tested whether functionally bilingual politicians systematically alter rhythm across languages in high-stakes settings, or how much rhythmic variance reflects speaker identity, language, and gender.

The present study addresses these gaps using spontaneous political speech from ten high-profile Luxembourgish politicians who regularly speak Luxembourgish and French in public. The material comes from the same bilingual corpus as our previous work on voice quality and global prosodic correlates of charismatic speech in this speaker group, but here we focus exclusively on rhythmic organization. For each speaker, we extract 20 spontaneous sentences in Luxembourgish and 20 in French from comparable public settings, annotate consonant-vowel structure and pauses, and derive duration-based rhythm metrics from these annotations. Manzoni Luxenburger~\cite{manzoni2021intonation} and Thill~\cite{thill2017etude} together provide a detailed account of Luxembourgish intonation, vowel quality, and duration that informs expectations about \%V and vowel expansion in bilingual speech. Against this empirical background, we address three questions.

\textbf{RQ1. Speaker versus language in rhythm.} How much of the variance in duration-based rhythm metrics is due to stable differences between speakers compared to language choice? We estimate this using speaker ICC and language $R^2$ from aggregated speaker-by-language data.


\textbf{RQ2. Systematic Luxembourgish–French rhythm differences.} Do individual speakers systematically modulate durational patterns when switching between Luxembourgish and French in comparable political contexts? We test whether languages differ in the distribution of vocalic and consonantal durations, variability measures such as $\Delta C$, $\Delta V$, and rPVI, and sentence-level measures like pause counts and speaking rate.

\textbf{RQ3. Gender and bilingual rhythm strategies.} Do language-specific rhythm settings align for female and male politicians, or do gendered strategies emerge? We test this via a Language $\times$ Gender interaction in the models and inspect effect sizes for key metrics.

Our goal is not to infer perceived charisma directly, but to describe how bilingual speakers structure rhythmic timing across two languages that occupy distinct structural, phonological, phonotactic, and sociolinguistic roles.

\section{Methods}

\subsection{Speakers and material}

We analyze speech from ten bilingual Luxembourgish politicians (5 female, 5 male) who produced comparable political discourse in both Luxembourgish and French~\footnote{The data was collected from RTL Archive with permission.}. Rhythm metrics were extracted from speech files from parliamentary recordings, press conferences, and interviews. For each speaker, we selected 20 spontaneous sentences in Luxembourgish and 20 in French from comparable communicative settings, yielding a total of 400 sentence tokens. Sentences were defined as intonationally and syntactically coherent units. They were segmented manually in Praat based on audible prosodic boundaries, supported by punctuation in transcripts where available. All speech is spontaneous, not read. Segments with overlapping talk, strong background noise, or prominent non-speech vocalizations such as laughter or coughing were excluded.

\subsection{Segmentation and CV tier annotation}

Audio was extracted from source videos as 16-kHz mono signals. Sentence boundaries were refined in Praat. For each sentence, we created a consonant-vowel (CV) tier marking every segmental interval as consonantal or vocalic. WebMAUS provided initial language-specific alignments~\cite{kisler2017multilingual}, which were manually corrected in Praat using waveform and spectrogram evidence and identical C/V criteria across languages~\cite{pollak2008phone}. Silences $\geq 200$~ms were labeled as pauses, while stop and affricate closures remained consonantal intervals~\cite{machavc2009principles}.

The resulting TextGrid files include a sentence tier and a CV tier with alternating C and V intervals. From the CV tier, we computed four interval sequences for each sentence:
\begin{enumerate*}[label=\roman*), itemjoin={{, }}, itemjoin*={{, and }}]
  \item vocalic intervals (V)
  \item consonantal intervals (C)
  \item voiced intervals
  \item silent pauses
\end{enumerate*}.


\subsection{Rhythm metrics}

Our rhythm analysis follows~\cite{niebuhr2024rhythm}. From the CV and pause tiers, we derived duration-based measures for each sentence. We computed mean and standard deviation of consonant and vowel durations ($\overline{C}$, $\overline{V}$, $\Delta C$, $\Delta V$), raw and normalized pairwise variability indices (rPVI, nPVI) for consonantal and vocalic intervals, mean and standard deviation of syllable and intensity peak durations, and sentence level speaking rate in syllables and segments per second (Rate Syl, Rate Seg). All duration measures were computed in milliseconds. Distributions were inspected for outliers and skewness, skewed metrics such as pause counts and rPVI values were log transformed, and all metrics were then $z$ scored across the full dataset (mean 0, standard deviation 1) prior to modeling so that fixed effect estimates can be interpreted as standard deviation differences between conditions.
\subsection{Statistical analysis}

Whereas~\cite{niebuhr2024rhythm} contrasted neutral and charismatic styles using MANCOVA, our bilingual data comprise files per speaker and language in ecologically valid political contexts. We aggregated rhythm metrics across files to obtain one value per speaker per language, yielding a balanced within-speaker design.

\textbf{Variance decomposition (RQ1).} For each rhythm metric, we partitioned variance into components attributable to speaker identity versus language choice. Speaker-specific variance was quantified with the intraclass correlation coefficient (ICC) from a one-way random effects model. Language-driven variance was quantified as R² from a one-way ANOVA by language. 

\textbf{Within-speaker language effects (RQ2).} To test whether individual speakers systematically modulate durational patterns when switching languages, we used paired t-tests comparing Luxembourgish and French values within each speaker. Effect sizes were computed as Cohen's d (paired) to quantify the magnitude of within-speaker modulation.

\textbf{Language × Gender interaction (RQ3).} We tested whether language rhythm patterns differed by gender using two-way ANOVA models with Language, Gender, and their interaction as factors. We report p-values for the Language × Gender interaction term. The P-values for language main effects and interactions were adjusted across the 23 rhythm metrics using Benjamini--Hochberg false discovery rate correction to control the expected proportion of false discoveries. Statistical analyses were performed in Python using scipy.stats for paired t-tests, statsmodels for ANOVA models, and scikit-learn for variance decomposition. With N = 10 speakers providing paired observations, our design had 80\% power to detect large effects (Cohen's d $\geq$ 0.9) at $\alpha = 0.05$ across the 23 rhythm metrics, including two speaking rate measures (Rate Syl, Rate Seg).

\section{Results}

\begin{figure}[hpt!]
    \centering
    \includegraphics[width=\linewidth]{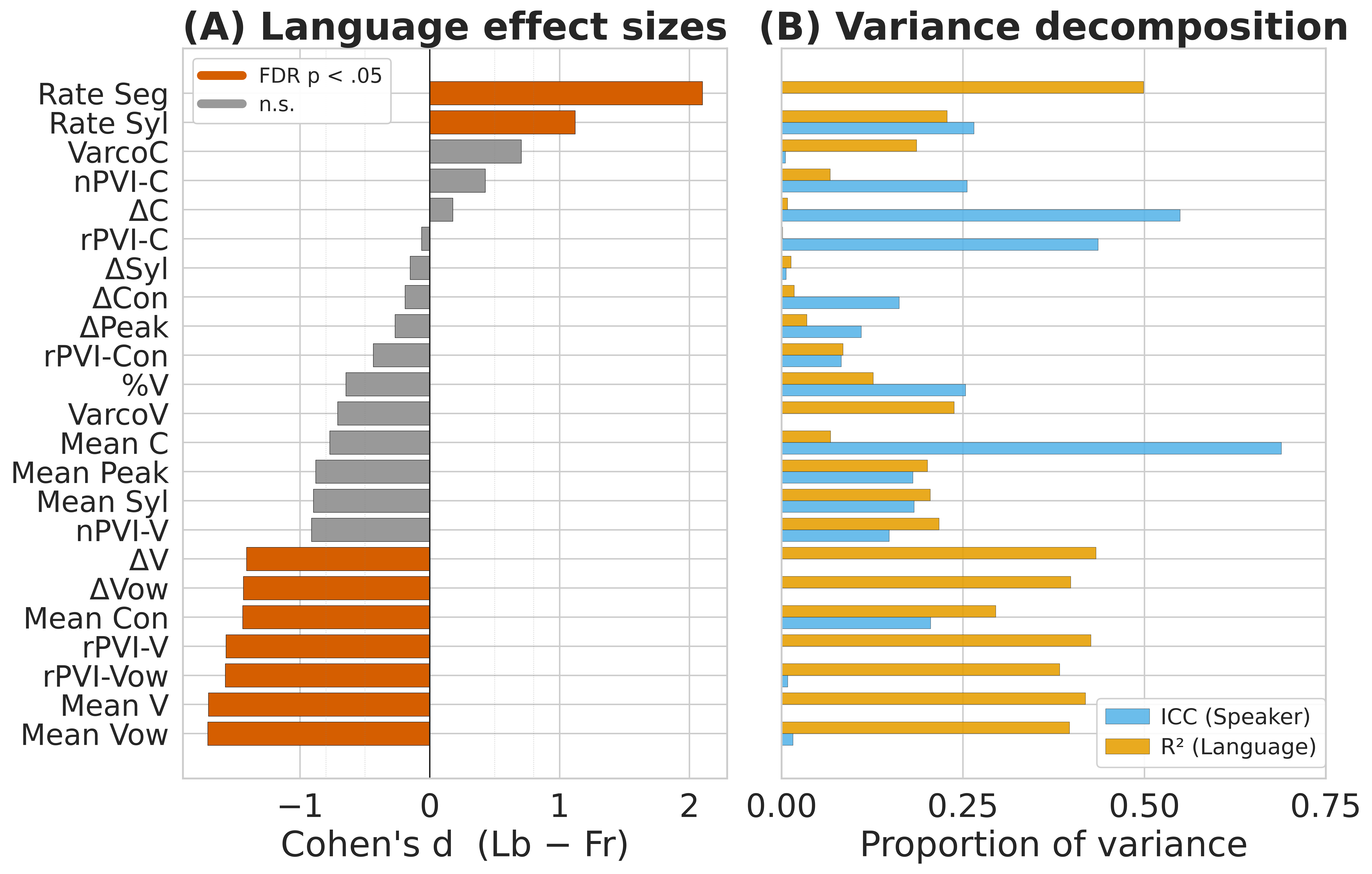}
    \caption{Rhythm metrics. (A) Forest plot of Cohen's d (Lb - Fr). (B) Variance decomposition (ICC speaker, R$^2$ language).}
    \label{fig:forest}
\end{figure}

Table~\ref{tab:rhythm_full} summarizes means, paired effect sizes,
and variance components, while Figures~\ref{fig:forest} and
\ref{fig:radar} visualize language effects, variance decomposition,
and speaker-specific rhythmic profiles.

\begin{table}[hpt!]
    \centering
    \caption{Metrics for Luxembourgish (Lb) and French (Fr). $p_{\text{Lang}}$ and $p_{\text{L}\times\text{G}}$ are Benjamini--Hochberg FDR-adjusted $p$ values.}
    \label{tab:rhythm_full}
    \scriptsize
    \setlength{\tabcolsep}{4pt}%
    \begingroup
    \renewcommand{\arraystretch}{1}
    \resizebox{\columnwidth}{!}{%
    \begin{tabular}{l r r r r c r r r}
        \toprule
        Metric & Lb & Fr & $d$ &
        $p_{\text{Lang}}$ & Sig. &
        ICC & $R^2_{\text{Lang}}$ &
        $p_{\text{L}\times\text{G}}$ \\
        \midrule
        \multicolumn{9}{l}{\textit{Segmental C.V metrics}} \\
        Mean C   & 0.0994 & 0.1043 & -0.77 & 0.0889 &       & 0.689 & 0.068 & --    \\
        Mean V   & \hlLux{0.0886} & \hlFr{0.1172} & -1.71 & 0.0049 & ***   & 0.000 & 0.419 & --    \\
        $\Delta$C   & 0.0622 & 0.0608 & +0.18 & 0.6995 &       & 0.549 & 0.008 & 0.671 \\
        $\Delta$V   & \hlLux{0.0607} & \hlFr{0.0950} & -1.41 & 0.0059 & ***   & 0.000 & 0.433 & 0.441 \\
        VarcoC   & 0.6267 & 0.5815 & +0.70 & 0.0958 &       & 0.005 & 0.186 & 0.782 \\
        VarcoV   & 0.6817 & 0.7990 & -0.71 & 0.0958 &       & 0.000 & 0.238 & 0.441 \\
        rPVI.C   & 6.0854 & 6.1348 & -0.06 & 0.8887 &       & 0.436 & 0.001 & 0.671 \\
        rPVI.V   & \hlLux{5.5494} & \hlFr{8.2633} & -1.57 & 0.0049 & ***   & 0.000 & 0.426 & 0.441 \\
        nPVI.C   & 58.8818 & 56.9506 & +0.43 & 0.2842 &       & 0.256 & 0.067 & 0.671 \\
        nPVI.V   & 55.6212 & 59.8615 & -0.91 & 0.0571 &       & 0.148 & 0.217 & 0.441 \\
        \midrule
        \multicolumn{9}{l}{\textit{Consonant vs vowel metrics}} \\
        Mean Con & \hlLux{0.0718} & \hlFr{0.0831} & -1.44 & 0.0059 & ***   & 0.205 & 0.295 & --    \\
        Mean Vow & \hlLux{0.0811} & \hlFr{0.1052} & -1.71 & 0.0049 & ***   & 0.016 & 0.397 & --    \\
        $\Delta$Con & 0.0411 & 0.0431 & -0.19 & 0.6995 &       & 0.162 & 0.017 & 0.984 \\
        $\Delta$Vow & \hlLux{0.0511} & \hlFr{0.0767} & -1.44 & 0.0059 & ***   & 0.000 & 0.398 & 0.441 \\
        rPVI.Con & 4.0060 & 4.4475 & -0.44 & 0.2842 &       & 0.082 & 0.085 & 0.984 \\
        rPVI.Vow & \hlLux{4.6933} & \hlFr{6.6979} & -1.58 & 0.0049 & ***   & 0.008 & 0.383 & 0.441 \\
        \midrule
        \multicolumn{9}{l}{\textit{Syllable.level metrics}} \\
        Mean Syl & 0.1950 & 0.2234 & -0.90 & 0.0571 &       & 0.183 & 0.205 & 0.894 \\
        $\Delta$Syl & 0.1510 & 0.1645 & -0.15 & 0.7380 &       & 0.006 & 0.013 & 0.752 \\
        Mean Peak & 0.2090 & 0.2423 & -0.88 & 0.0580 &       & 0.181 & 0.201 & --    \\
        $\Delta$Peak & 0.2009 & 0.2266 & -0.27 & 0.5473 &       & 0.110 & 0.035 & 0.984 \\
        \midrule
        \multicolumn{9}{l}{\textit{Speaking rate}} \\
        Rate Syl & \hlLux{5.2107} & \hlFr{4.5586} & +1.12 & 0.0204 & *     & 0.265 & 0.228 & 0.885 \\
        Rate Seg & \hlLux{12.3842} & \hlFr{9.6180} & +2.10 & 0.0020 & **    & 0.000 & 0.499 & 0.885 \\
        \bottomrule
    \end{tabular}%
    }
    \endgroup
    \tiny
    Sig.: * $p_{\text{Lang}}<.05$, ** $p_{\text{Lang}}<.01$, *** $p_{\text{Lang}}<.001$ \tiny {(FDR.corrected)}
\end{table}

\subsection{Language and segmental and syllable metrics (RQ2)}

Across segmental C and V measures, clear language effects appear mainly for vowel-based metrics. Mean vowel duration and its variability are higher in French than in Luxembourgish, with large paired effect sizes for Mean V and $\Delta$V (Cohen's $d=-1.71$ and $-1.41$, respectively, both FDR corrected $p<.01$). The raw pairwise variability index for vowels behaves similarly, with higher rPVI.V in French (Cohen's $d=-1.57$, FDR corrected $p<.01$). For the same speakers in comparable political contexts, vowels are therefore longer and more variable when politicians speak French. In contrast, consonant metrics show much weaker or non-significant language differences. Mean C, $\Delta$C, VarcoC, rPVI.C, and nPVI.C reach at most medium effect sizes, and none survive FDR correction. This asymmetry is visible in Figure~\ref{fig:forest}, where vowel-based metrics cluster toward sizeable negative d values (Lb $<$ Fr), while consonant-based metrics remain close to zero.

Derived consonant versus vowel interval metrics reinforce this pattern. French shows longer and more variable vocalic intervals (Mean Vow, $\Delta$Vow, rPVI.Vow, all $|d|\geq 1.44$ with FDR corrected $p<.01$), whereas consonant interval metrics such as Mean Con and rPVI.Con exhibit weaker, non-significant shifts. The proportion of vocalic material (\%V, not shown in Table~\ref{tab:rhythm_full}) trends higher in French (Cohen's $d\approx -0.65$, FDR corrected $p=0.11$), consistent with a more vowel-dense timing pattern. Syllable level measures show changes in the same direction, with moderate negative $d$ values and FDR corrected $p$ values around 0.06, suggesting syllable duration differences that are smaller than the segmental vowel effects.

Speaking rate metrics show equally strong language effects. Luxembourgish tokens are produced at significantly higher rates than French tokens for both syllable rate (Rate Syl: $d=+1.12$, FDR corrected $p=.020$) and segment rate (Rate Seg: $d=+2.10$, FDR corrected $p=.002$; Table~\ref{tab:rhythm_full}). Exploratory consonant and vowel-specific rate measures show the same pattern for vowels, but not for consonants, so we report only overall syllable and segment rate in the main table. These findings confirm that politicians speak substantially faster in Luxembourgish than in French when addressing comparable audiences.

\subsection{Speaker versus language contributions (RQ1)}

Variance decomposition clarifies how much of the observed variation is driven by stable speaker differences versus language choice. For the key vowel-based metrics (Mean V, $\Delta$V, rPVI.V, Mean Vow, $\Delta$Vow, rPVI.Vow), language accounts for a substantial share of the variance, with $R^2_{\text{Lang}}$ values between 0.38 and 0.43, while intraclass correlations for speaker identity are near zero. The same profile is observed for speaking rate, particularly Rate Seg, where language explains roughly half of the variance ($R^2_{\text{Lang}} \approx 0.50$) and the ICC for speaker identity is near zero. Thus, cross-language shifts dominate over stable between-speaker clustering for vowels, vocalic intervals, and speaking rate. By contrast, several consonant-based metrics show higher ICC values and much lower $R^2_{\text{Lang}}$, suggesting that consonant timing carries relatively strong speaker-specific signatures and only weak language effects. Syllable-level metrics occupy an intermediate position, with both ICC and $R^2_{\text{Lang}}$ in the low-to-mid range. Together, these results partly confirm RQ1. Rhythm metrics do capture both speaker- and language-specific structure, but the balance between them depends on the metric family. Vowels and vocalic intervals are more clearly language-driven, while consonantal and some syllable-level measures preserve more speaker-linked variability.

\subsection{Per speaker rhythmic fingerprints}

\begin{figure}[hpt!]
    \centering
    \includegraphics[width=\linewidth]{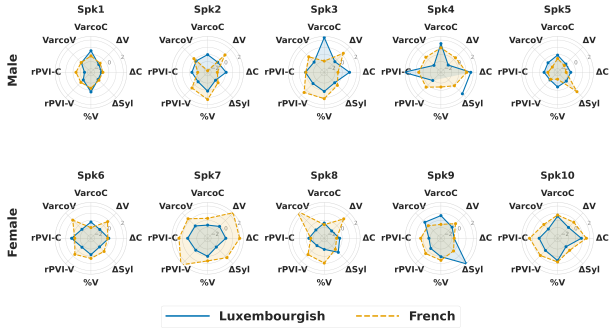}
    \caption{Rhythmic fingerprints of ten bilingual politicians. Radar plots show z-scored duration-based rhythm metrics for Luxembourgish (solid blue) and French (dashed red).} 
    \label{fig:radar}
\end{figure}

The radar plots in Figure~\ref{fig:radar} visualize z-scored rhythm profiles for each politician across core metrics. Within speakers, the shapes of the Luxembourgish and French contours are similar, so compact profiles in Luxembourgish tend to remain compact in French and expanded profiles remain expanded, supporting stable rhythmic “fingerprints”. At the same time, French contours show a consistent outward shift along vowel-related axes such as $\Delta$V and rPVI.V, whereas consonant-related axes such as $\Delta$C and rPVI.C largely overlap across languages. The magnitude of these shifts varies across speakers, with politicians showing pronounced French expansions in the vowel domain and others displaying tighter overlap, which matches the group-level statistics in Table~\ref{tab:rhythm_full} where vowel timing emerges as the main locus of language-dependent adjustment and consonant timing preserves more of the individual rhythmic fingerprint.

\subsection{Gender and rhythm strategies (RQ3)}

Language by Gender interactions were tested for all metrics using two-way ANOVA and Benjamini--Hochberg correction across the family of tests. As shown in Table~\ref{tab:rhythm_full}, almost all $p_{\text{L}\times\text{G}}$ values are non-significant, and no systematic pattern emerges in the small numerical differences that remain. Direction and magnitude of the Luxembourgish vs. French rhythmic contrasts are similar for female and male politicians across metric families, including speaking rate. So, in this sample of high-profile speakers, women and men appear to implement comparable bilingual rhythm strategies. Politicians modulate vowel timing and speaking rate when switching from Luxembourgish to French, but does not differ as a matter of gender.

\section{Discussion and conclusions}

This study disentangles speaker- and language-specific contributions to duration-based rhythm metrics in bilingual political speech. Using spontaneous speech from ten politicians who regularly switch between Luxembourgish and French in comparable public contexts, we asked whether rhythm metrics primarily index stable speaker fingerprints (RQ1), whether speakers systematically modulate durational patterns across languages (RQ2), and whether such modulation differs by gender (RQ3).

Overall, the results show a clear asymmetry between consonantal and vocalic metrics. Consonant-related measures such as Mean C, $\Delta$C, VarcoC, and rPVI.C show moderate language effects, with several such metrics showing high intraclass correlations (ICC) and small $R^2_{\text{Lang}}$ values (Table~\ref{tab:rhythm_full}). By contrast, vowel-related measures show large systematic language differences. Mean vowel duration, its variability ($\Delta$V), and the rPVI for vowels are all substantially higher in French than in Luxembourgish, with large within-speaker effect sizes and FDR-corrected $p$ values below .01. The same pattern holds for vocalic interval measures (Mean Vow, $\Delta$Vow, rPVI.Vow), which show strong temporal expansions for French relative to Luxembourgish. Speaking rate shows an equally strong language effect, with Luxembourgish being produced at faster rates than French (syllable rate $d=+1.12$; segment rate $d=+2.10$), and variance dominated by language rather than speaker identity (for Rate Seg, $R^2_{\text{Lang}} \approx 0.50$). These findings support RQ2. Bilingual politicians do not keep rhythm metrics constant across languages. Instead, vocalic timing and overall tempo are systematically reconfigured for Luxembourgish and French~\cite{grabe2026durational,white2007calibrating}.

Variance decomposition clarifies the roles of speaker identity and language choice. For key vocalic metrics, $R^2_{\text{Lang}}$ lies in the range 0.38–0.43, while ICC values are close to zero, indicating that cross-language shifts dominate over between-speaker clustering. Speaking rate, especially Rate Seg, follows the same pattern. Consonantal metrics show the opposite profile, with higher ICC and lower $R^2_{\text{Lang}}$, suggesting that consonant timing is a domain where individual speakers' idiosyncratic temporal profiles are less constrained by language. Together, this partly supports RQ1. Evidence emerges for both speaker-specific rhythmic signatures and language-dependent rhythm settings; the balance between them depends on the metric family~\cite{dellwo2015rhythmic}.

No Language × Gender interactions emerged after FDR correction. For almost all metrics, $p_{\text{L}\times\text{G}}$ values were clearly non-significant, i.e. the Luxembourgish–French contrast was implemented similarly for women and men, including speaking rate. So, in terms of RQ3, within this sample of high-profile politicians, female and male speakers used comparable rhythm patterns when switching languages, even though segment durations and global tempo differ substantially across languages~\cite{niebuhr2024rhythm}. In light of the charisma gender gap documented for other acoustic melodic cues~\cite{novak2017gender,voss2024beautiful}, the results suggest that duration-based rhythm metrics function more as a gender-neutral backdrop than as a primary resource for closing this gap.

Our study design also includes limitations. The sample size is modest, and all speakers are elite politicians, so the results probably characterize a specific communicative niche rather than general bilingual behavior. To obtain a balanced within-speaker design, rhythm metrics were aggregated at the speaker-by-language level, yielding paired comparisons but collapsing sentence-level variability and hence reducing degrees of freedom for variance partitioning and interaction tests. The study also focuses on production and does not test whether the observed rhythm differences are perceptually salient and/or contribute to the speakers' perceived charisma.

Despite these caveats, the findings demonstrate that duration-based rhythm metrics reveal systematic language-dependent timing strategies in naturalistic, high-stakes bilingual speech. They extend previous work on monolingual charismatic rhythm by showing that, for experienced bilingual speakers, rhythm is not only a resource for stylistic variation within a language but also a domain in which language choice itself shapes the temporal organization of speech~\cite{white2007calibrating}. In Luxembourg's multilingual political arena, speech rhythm therefore reflects language-specific temporal organization while remaining largely gender-neutral.


\section{Acknowledgments}
This research was conducted in the context of the LuxVoice
project, funded by the Luxembourg National Research Fund
(FNR) under grant agreement (project reference 19205922).
LuxVoice aims to advance Luxembourgish language technologies and support the development of robust multilingual
AI resources. The authors gratefully acknowledge the support of RTL Lëtzebuerg and the University of Luxembourg.

\section{Generative AI Use Disclosure}
The use of generative AI (i.e., Grammarly) in this paper was limited strictly to text refinement and clarity improvements. The authors are solely responsible for all scientific content, including the methodology, experiments, analysis, and conclusions.

\bibliographystyle{IEEEtran}
\bibliography{mybib}

\end{document}